%% file: main.tex
\documentclass{article}

\usepackage[accepted]{sysml2019}
\usepackage{microtype}
\usepackage{mathptmx} 
\usepackage{graphicx}
\usepackage[table,xcdraw]{xcolor}
\newcommand{\ignore}[1]{}
\usepackage{fancyhdr}
\usepackage[normalem]{ulem}
\usepackage[hyphens]{url}
\usepackage{microtype}
\usepackage{algorithmic}
\usepackage{fancyhdr}
\usepackage{array}
\usepackage{fixltx2e}
\usepackage{stfloats}
\usepackage{url}
\usepackage{times}
\usepackage{color}
\usepackage{verbatim}
\usepackage{hyperref}
\usepackage{subfiles}
\usepackage{caption}
\usepackage{subcaption}
\usepackage{lipsum}
\usepackage{xr}
\usepackage{nameref}
\usepackage{zref-xr}
\usepackage{adjustbox}
\usepackage{amsmath, amssymb}
\usepackage[ansinew]{inputenc}
\usepackage{mathtools}
\usepackage{esvect}
\usepackage{xspace}
\usepackage{standalone}
\usepackage{fixltx2e}
\usepackage{authblk}
\usepackage{xcolor}
\usepackage{booktabs}
\usepackage{multirow}
\usepackage{cleveref}
\usepackage{tabularx}
\usepackage{gensymb}
\usepackage{amsmath,stmaryrd,pict2e,picture}
\usepackage{siunitx}
\usepackage{grffile}

\usepackage{hyperref}

\newcommand{\autop}[0]{AutoSoC\xspace}
\newcommand{\Fig}[1]{Fig.~\ref{#1}}

\makeatletter
\newcommand{\pictslash}[2]{%
  \vcenter{\hbox{%
    \sbox0{$\m@th#1\varobslash$}\dimen0=.55\wd0
    \pictslash@aux#2%
  }}%
}
\newcommand{\pictslash@aux}[2]{%
    \begin{picture}(\dimen0,\dimen0)
    \roundcap
    \put(0,#1\dimen0){\line(1,#2){\dimen0}}
    \end{picture}%
}
\sysmltitlerunning{AutoSoC: Automating Algorithm-SOC Co-design for Aerial Robots}

\begin{document}

\twocolumn[
\sysmltitle{AutoSoC: Automating Algorithm-SOC Co-design for Aerial Robots}




\begin{sysmlauthorlist}
\sysmlauthor{Srivatsan Krishnan}{0}
\sysmlauthor{Thierry Tambe}{0}
\sysmlauthor{Zishen Wan}{0}
\sysmlauthor{Vijay Janapa Reddi}{0}
\sysmlaffiliation{0}{John A. Paulson School of Engineering, Harvard University, Cambridge, MA, USA}

\sysmlcorrespondingauthor{Srivatsan Krishnan}{srivatsan@seas.harvard.edu}
\end{sysmlauthorlist}

\centering \textsuperscript{1}Harvard University \\ Contact: srivatsan@seas.harvard.edu

\sysmlkeywords{Machine Learning, SysML}

\vskip 0.3in

\input{abstract.tex}]
\input{intro.tex}
\input{background.tex}
\input{autox.tex}
\input{eval.tex}
\input{results.tex}
\input{conclusion.tex}



 \printAffiliationsAndNotice{}  

\nocite{langley00}

\bibliography{ref}
\bibliographystyle{sysml2019}


\end{document}

%% file: abstract.tex
\begin{abstract}
Aerial autonomous machines (Drones) has a plethora of promising applications and use cases. While the popularity of these autonomous machines continues to grow, there are many challenges, such as endurance and agility, that could hinder the practical deployment of these machines. The closed-loop control frequency must be high to achieve high agility. However, given the resource-constrained nature of the aerial robot, achieving high control loop frequency is hugely challenging and requires careful co-design of algorithm and onboard computer. Such an effort requires infrastructures that bridge various domains, namely robotics, machine learning, and system architecture design. To that end, we present AutoSoC, a framework for co-designing algorithms as well as hardware accelerator systems for end-to-end learning-based aerial autonomous machines. We demonstrate the efficacy of the framework by training an obstacle avoidance algorithm for aerial robots to navigate in a densely cluttered environment. For the best performing algorithm, our framework generates various accelerator design candidates with varying performance, area, and power consumption. The framework also runs the ASIC flow of place and route and generates a layout of the floor-planed accelerator, which can be used to tape-out the final hardware chip.  
\end{abstract}

%% file: intro.tex
\section{Introduction}
\label{sec:intro}
Autonomous machines are increasingly playing a key role in several industries, such as transportation~\cite{Timothy2017}, medical care~\cite{medical-org}, agriculture~\cite{8373043}, mining~\cite{mining} etc. The autonomous machine is a board term that encompasses several classes of robots, such as a self-driving car, a robot arm, aerial robot, etc. An aerial robot such as quadcopter is a special category of autonomous machines that has several unique capabilities such as the ability to vertical take-off and land (VTOL), ability to navigate in confined spaces, among many others. These capabilities allow them to be deployed in several applications such as search and rescue~\cite{search-and-rescue,search-and-rescue-org}, package delivery~\cite{package-org-1,package-org-2}, aerial survey~\cite{survey-org}, surveillance~\cite{surveillance-org}, sports photography~\cite{photographyh-org}, entertainment~\cite{entertainment-org,droneshow}.

In spite of the promising applications of aerial robots, there are a couple of challenges that restrict them from achieving full potential. First, aerial robots are mobile and have a limited onboard battery capacity, which is used to power the onboard electronics and as well as the rotors. For instance, many commercial drones that are used for the applications mentioned above, the maximum flight time on a fully charged battery varies anywhere between 6 mins to 20 mins~\cite{6mins,20mins}. In the scenario where these are deployed for mission-critical applications, limited flight time severely impacts the quality of service. For instance, for package delivery missions, a 15 mins flight time severely restricts the maximum range where package delivery service can be deployed.  

\begin{figure}[t]
        \includegraphics[width=1.0\columnwidth]{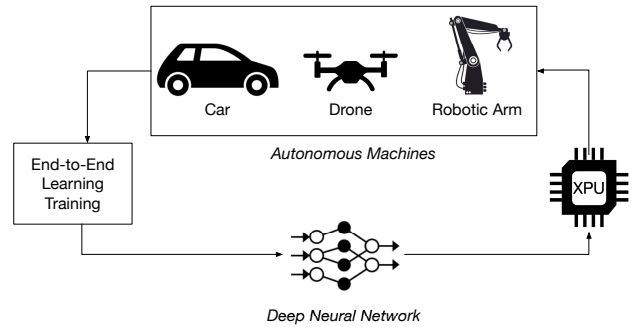}
        \caption{End-to-End learning in autonomous machines.} 
       \vspace{-20pt} \label{fig:e2e-flow}
\end{figure}

Prior studies have shown that compute latency plays a role in increasing the speed of the aerial robot, which can result in saving energy by finishing the mission faster~\cite{Bor2018}. The target of minimizing the compute latency makes it interesting for architects to design power-performance efficient specialized hardware accelerators for these emerging domains. Until now, there has been only limited research in this area, such as PULP-Shield~\cite{pulp-dronet}, which uses custom hardware accelerators for aerial robot navigation tasks. However, these hardware accelerators were at best point solution, which was explicitly targeted for nano-drones and one particular algorithm.


The compute latency depends upon the choice of the algorithm used for aerial robot navigation. Traditionally, the algorithms used for controlling the aerial robot are based on the sense-plan-act paradigm~\cite{octomap,slam,vins-mono,a-star,prm,rrt}. The overall compute latency for algorithms that use a sense-plan-act paradigm is typically in the order of a few seconds~\cite{high-speed-ppc}, which determines the time taken to react to a change in sensor input. In contrast, the emerging algorithmic paradigm called End-to-End (E2E) learning provides significant promise in replacing traditional sense-plan-act with a single neural network (E2E model)~\cite{cad2rl,dronet,learn2crash,trailnet,trailnet} as shown in \Fig{fig:e2e-flow}. In E2E learning, we train an E2E model for a particular robot. Once the trained model achieves sufficient algorithmic performance, it is deployed onto the onboard compute platform in the robot.

However, unlike computer vision tasks such as image recognition, object detection, E2E learning for aerial robots lacks standardization. The E2E model architecture is hand-crafted depending upon the robot task~\cite{dronet,trailnet}, and the availability of data. Hence this affects the scalability and practical deployment or E2E model for the aerial robots. Hence there is a need to develop an infrastructure that allows us to quickly iterate over the design of E2E models quickly and efficiently. Also, the infrastructure should be capable of generating efficient hardware accelerators for the E2E models to minimize the processing latency to increase the closed-loop control frequency.

\begin{figure}
        \includegraphics[width=\columnwidth]{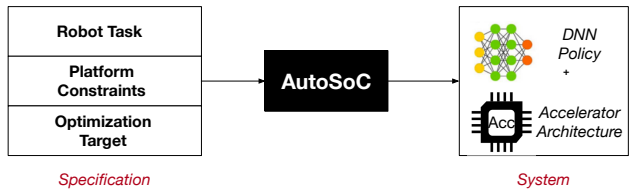}
        \caption{High-level flow in AutoSoC framework.}
        \vspace{-20pt}
     \label{fig:autopilot-high-level}
\end{figure}

To that end, we introduce a new comprehensive framework called AutoSoC (\Fig{fig:autopilot-high-level}) that allows us to perform algorithm-hardware co-design for a given task for the aerial robot. The AutoSoC framework co-designs both the E2E model along with the hardware accelerator to meet the final domain-specific optimization targets, such as to minimize the energy of flight and maximize the success rate for a given environment.

\Fig{fig:autopilot-high-level} shows a high-level overview of the \autop framework. The framework takes a user-defined specification, which comprises three components, namely : (i) robot's task information (e.g., autonomous navigation, number of obstacles); (ii ) Platform constraints such as the type of sensor and compute used; (iii)  domain-specific optimization targets such as minimize energy of flight while achieving success rate above a given threshold. Using these specifications, \autop under the hood, uses two distinct tools: (i) Air Learning~\cite{krishnan2019} to simulate an E2E neural network model to determine its impact on quality-of-flight (e.g., its success rate) for a given task; (ii) FLexACL to model and simulate a deep neural network-based hardware accelerator. Using these two tools allows us to perform a comprehensive design space exploration by simultaneously tuning the parameters for both E2E models as well as accelerator's micro-architecture to yield different accelerator candidates that achieve the target specified in the high-level specification.

Using our proposed framework, we determine an E2E model and accelerator designs for aerial robot navigation in a cluttered environment. The E2E model is chosen such that it has a maximum success rate and high performance (lower compute latency). Using the framework, we design a point-to-point navigation policy capable of navigating in a cluttered environment. The policy achieves a success rate of 91\%. We also generate six different accelerator candidates whose processing times vary from as low as 4.8 microseconds to 60.2 microseconds. The power envelope of the placed-and-routed accelerators varies from 0.142 W to 1.091 W. Likewise, the area of the design varies from 4.9~mm\textsuperscript{2} to 39.20~mm\textsuperscript{2}. Unlike prior work such as DroneNet~\cite{pulp-dronet,dronet}, SOC accelerators generated from the \autop framework is generalized and can generate E2E-model/SOC accelerators for a nano-sized aerial robot to standard size aerial robots.

In summary,  we make the following contributions:
\begin{itemize}
    \item {We introduce AutoSoC, a comprehensive framework that allows us to co-design algorithm as well as system on a chip (SOC) for a given aerial robot task.}
   \item {For a given robot task, using AutoSoC, we design an efficient E2E model that achieves a 91\% success rate and also generates a variety of different accelerator design candidates for processing the E2E model. }
    \item {We generate the accelerator chip layout for the accelerator candidate that has the best performance/power and lowest area, thus able to explore the design space starting from robot task to SOC chip design.}
\end{itemize}

The rest of the paper is organized as follows. Section~\ref{sec:background} provides the background on the aerial autonomous machine and different types of  E2E learning methods. Section~\ref{sec:autox} introduces various components of \autop framework in detail. Section~\ref{sec:eval}, we discuss our evaluation methodology and the different parameters used in our experiments. Section~\ref{sec:results} present the exhaustive experimental analysis and results to show the effectiveness of our \autop framework in determining optimal E2E models and accelerator designs. Section~\ref{sec:conclusion} concludes the paper and goes over the future directions for this work.

%% file: background.tex
\section{Background}
\label{sec:background}
In this section, we provide background on the diversity of autonomous aerial machines and their taxonomy in terms of their size, weight, and power consumption. Next, we provide a brief background of two commonly used end-to-end learning methods. Then we provide a background on how end-to-end learning is applied in the context of aerial robots and provide definitions for the metrics we use in the rest of the paper. Lastly, we provide a background on hardware accelerator effort for processing deep neural networks efficiently.
\subsection{Aerial Autonomous Machines}
Aerial autonomous machines are very diverse and come in different shapes, sizes, and performance under which they operate. Here we provide a comparison based on their size, weight, power, and onboard device. Table~\ref{tab:aerial-robot-background} tabulates the rotorcraft UAVs taxonomy by vehicle class-size and the range of values for weight and power~\cite{DBLP:journals/corr/abs-1805-01831}. On one end of the spectrum we have the standard-size aerial robot weighs about 1 Kg and has a power envelope of 100 W whereas on the other end, we have pico-sized drone which weighs 1000\textsuperscript{th} of the standard-size drone and as power envelope of less than 100~mW. Depending upon the power envelope, the type and capability of the onboard compute platform varies.

\begin{table}[]
\centering
\Large
\resizebox{\columnwidth}{!}{%
\begin{tabular}{|c|c|c|c|l|}
\hline
\textbf{Vehicle Class} & \textbf{$\varoslash$ : Weight [cm:kg]} & \textbf{Power [W]} & \textbf{Onboard Device} & \textbf{References} \\ \hline
\textbf{std-size} & $\sim$ 50 : $\geq$ 1 & $\geq$ 100 & Desktop & \begin{tabular}[c]{@{}l@{}} ~\cite{8119942}\\~\cite{typhoon}\\~\cite{parrot}\end{tabular} \\ \hline
\textbf{micro-size} & $\sim$ 25 : $\sim$ 0.5 & $\sim$ 50 & Embedded  & \begin{tabular}[c]{@{}l@{}}~\cite{Conroy:2009:IWI:1644233.1644234}\\~\cite{8588849}\\~\cite{falcore}\end{tabular}   \\ \hline
\textbf{nano-size} & $\sim$ 10 : $\sim$ 0.01 & $\sim$ 5 & MCU & \begin{tabular}[c]{@{}l@{}}~\cite{7833065} \\~\cite{holy}\\~\cite{spark}\end{tabular} \\ \hline
\textbf{pico-size} & $\sim$ 2 : $\leq$ 0.001 & $\sim$ 0.1 & ULP & \begin{tabular}[c]{@{}l@{}}~\cite{Wood:2012:PPA:2370704.2370708} \\~\cite{1}\end{tabular} \\ \hline
\end{tabular}%
}

\caption{Taxonomy of aerial robot based on their size~\cite{DBLP:journals/corr/abs-1805-01831}}
\label{tab:aerial-robot-background}
\vspace{-20pt}
\end{table}

\subsection{End-to-End Learning Methods}
End-to-End learning methods directly process input sensor information (such as RGB, Lidar, etc.) and produces output actions that are used to control the autonomous machines. Two popular end-to-end learning methods are typically used for sensorimotor control for autonomous machines are as follows:

\textbf{Supervised learning:}
One form of end-to-end learning can be formulated as supervised learning~\cite{e2e-nvidia,dronet,trailnet0,trail-net}. In this formulation, a human expert controls the autonomous machine (e.g., human driving a car), and his actions are recorded along with the sensor information. The sensor information is the data, and human action are the ground-truth labels for the data. Once sufficient data is collected, an E2E model for the autonomous machine is trained similar to supervised learning tasks such as image classification. Once the model achieves good accuracy, it is deployed to the robot.

One of the most significant advantages of end-to-end learning with supervised learning is access to expert action, which is typically the performance level one wants to achieve with autonomous machines. One of the shortcomings of the end-to-end learning approach is that the performance of the E2E model depends upon the quality and quantity of the data. Also collecting data for all autonomous machine might be logistically expensive. For instance, an aerial robot has only 20 min of flight time which severely limits the quantity of the data that can be collected.

\textbf{Reinforcement Learning:}
Reinforcement learning~\cite{rl-book} is another popular end-to-end learning technique that has also been successfully used for end-to-end control for several autonomous machines~\cite{cad2rl,qt-opt,rl-car}. Reinforcement learning (RL) is a form of self-supervised learning where the agent (robot) interacts with the environment and through trial and error, determines the best sequence of actions to maximize the long term reward. 
At every time step, the agent observes the current state and chooses an action. Because of the action, the agent moves in the environment and observes a new state. Along with the state transition, the agent gets a reward for the action it took in the previous state. If the action resulted in better progress towards the goal, the agent gets a positive reward. However, if the action results in an undesirable state, the agent is penalized. Using the reward, the agent optimizes its policy (E2E model), and once it has sufficient experiences, it learns the optimal policy to maximize the reward.

\textbf{DNN Inference Hardware Accelerators:} Over the last half-decade, there has been tremendous research efforts focused on improving the performance and energy efficiency of deep learning hardware accelerators~\cite{eie,minerva,diannao,eyeriss}. As these accelerators get deployed at all computing scales, from resource-constrained IoT devices to massive data center farms, there has been additional interest in the research community to auto-generate and customize them on-the-fly~\cite{magnet,scaledeep}. The FlexACL component of \autop provides accelerator generator capabilities to specialize in the compute of RL E2E policies on customized hardware satisfying a prescribed power-area-performance budget, and therefore optimizing the quality-of-flight metrics of aerial robots.   

%% file: autox.tex
\section{AutoSoC}
\label{sec:autox}
\begin{figure*}
        \includegraphics[width=2.0\columnwidth]{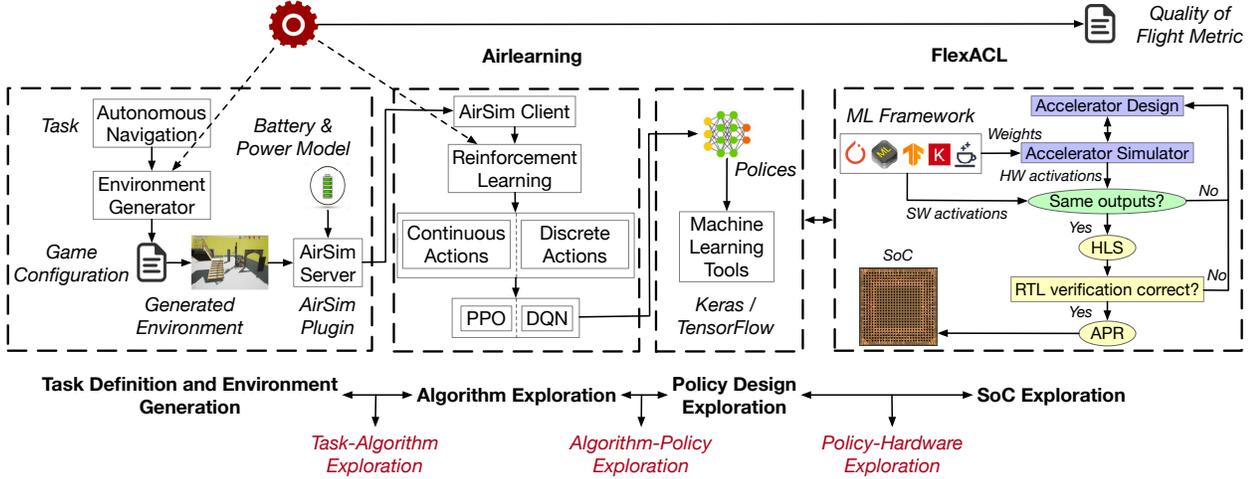}
        \caption{Different components in AutoSoC. In the front-end, we have Air Learning~\cite{airlearning}, which is used to perform task-algorithm-policy exploration. Once a policy is determined, it will be fed to the FlexACL block, which will take the neural network policy and generate a synthesizable RTL accelerator template to meet the performance and power specifications.}
       \vspace{-20pt} \label{fig:autox}
\end{figure*}

In this section, we describe the \autop framework in detail. \autop framework has two major components, namely Air Learning and FlexACL as shown in Figure~\ref{fig:autox}. Air Learning framework is used to design and validate the E2E model for a given robot task, and FlexACL is the back-end that uses HLS based flow to synthesize hardware accelerator for processing the E2E model efficiently.

\subsection{Air Learning}
\label{sec:airlearning-training}
\autop uses Air Learning~\cite{airlearning} as the robot simulator to train E2E models for aerial robot navigation. Air Learning provides an infrastructure with a configurable and random environment generator that can simulate a variety of challenging environments for the aerial robot navigation task. It also integrates stable-baseline~\cite{stable-baselines}, which provides a high-quality implementation of reinforcement learning algorithms that can be used to train E2E learning models for aerial robot navigation tasks. Air Learning uses Tensorflow~\cite{abadi2016tensorflow} as the back-end tool for training ML models.

Based on the specified robot task, the desired success rate, and another environment related specification, \autop launches several Air Learning training instances in parallel with different hyper-parameters for the E2E models and the parameters for Air Learning environment generator. The details on these parameters are described in Section~\ref{sec:airl-env-settings}.

The E2E models that achieve the required success rate (or other user-specified quality-of-flight metrics) are evaluated on a random environment to validate the task level functionality. The validated E2E models are then passed to the FlexACL framework, which takes the model definition and generates the final hardware SOC accelerator.

\subsection{FlexACL}\label{sec:flex-acl}

FlexACL is a modular accelerator template based on the SystemC+HLS flow. It generates a Verilog RTL from a SystemC/C++ source code producing a hardware accelerator with AXI slave interfaces~\cite{axi}, which can be plugged as an IP onto pre-defined SoC interfaces.

\begin{figure}[t]
    \centering
    \includegraphics[width=0.75\columnwidth]{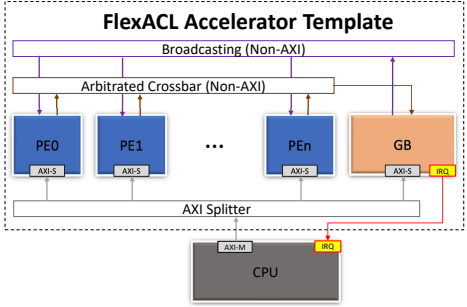}
    \caption{The FlexACL System. A CPU is used to send AXI configurations to the FlexACL accelerator which will then return back an interrupt signal upon completion of the layer computation.}
    \label{fig:flex_accel}
    \vspace{-15pt} 
\end{figure}

The architecture of the FlexACL accelerator template is shown in Figure~\ref{fig:flex_accel}. 
The communication between the accelerator's global buffer (GB) and processing elements (PEs) is performed via non-AXI channels. 
Notably, an arbiter is used to referee the stream of PE partial activation results, which will be aggregated by the GB.
Once the full activation has been collected, the GB will then broadcast it back to the PEs for the next layer computation.

A CPU with an AXI-Master interface is used to program and configure the target acceleration. Since there are only AXI-Slave ports inside the FlexACL accelerator template, we also implement an interrupt (IRQ) channel sent from the accelerator to CPU as an indication of completion of the computed task, which can be a fully-connected or RNN/LSTM/GRU layer. 

FlexACL instruction set architecture (ISA) allows the CPU to configure it with a particular neural network dimension (e.g., input and output size of FC layers) to create a customized acceleration pertaining to the generated E2E policy.

\begin{figure}[t]
    \centering
    \includegraphics[width=0.7\linewidth]{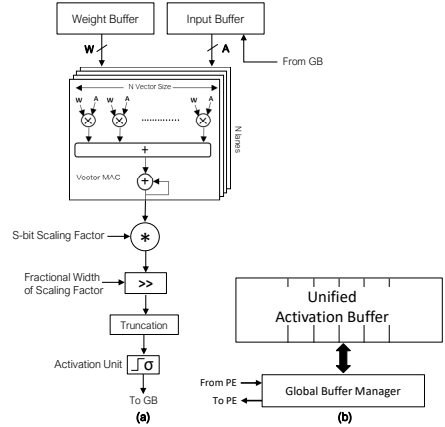}
    \caption{(a) Micro-architecture of the integer-based processing element (PE) and (b) global buffer (GB) of the FlexACL accelerator.}
    \label{fig:flex_pe_gb}
    \vspace{-10pt} 
\end{figure}

Figure~\ref{fig:flex_pe_gb} shows the micro-architecture of the PE and GB of the FlexACL accelerator. 
The PE contains $N$ fixed-point vector MAC units receiving $n-bit$ integer weight and activation vectors from their respective buffers. 
The MAC partial sums are stored in accumulation registers and then scaled by a high-precision scaling factor followed by a bit-shift to dequantize the computation~\cite{migacz2017}. Then, the data is clipped and truncated back to $n$ bits before being modulated by the neural network activation function.

The GB collects and unifies the partial activations computed by the PEs and then broadcast the complete activation back to each PE to process the next neural network layer. The GB contains a global buffer manager that generates the logical addresses for storing the input activations in the PE's weight buffer and the output activations in the GB's unified activation buffer. 

\subsection{Design Flow}
\label{sec:design-flow}

As shown in Figure~\ref{fig:autopilot-high-level}, Autopilot takes the following inputs: (i) robot tasks such as navigation, the target environment, and a threshold on the success rate;  (ii) E2E model hyper-parameters and (iii) an optimization target such as minimizing the accelerator power/area/runtime. The output of \autop is the optimal E2E NN policy and the corresponding accelerator architecture with the lowest energy metric. Below we present an overview of the two phases of the design flow.

In the first phase, a robot simulator is used to train various neural network policies for a given environment. The range of the NN policies are determined by the NN parameters that need to be tuned, for example, the number of layers and filters. For each possible NN policy, the simulator outputs the various quality-of-flight metrics, such as the success rate achieved when navigating the environment using the corresponding NN policy in the robot's onboard compute SoC. 
In AutoX, this simulator is Air Learning~\cite{krishnan2019}, which is an infrastructure to train E2E learning-based algorithms for aerial robots. Multiple instances of Air Learning can be used to train several E2E models candidates in parallel, for a given environment, and generate their respective success rates.  These E2E models that are below the threshold success rate are pruned out, and the E2E model that meets the criteria are passed to the next phase of the design flow. The details on the environment and other task-related settings used in Air Learning are presented in Section~\ref{sec:airlearning-training} below.

In the second phase, the E2E models that meet the success threshold criteria are selected and used as the input to the FlexACL framework. 
FlexACL closes the loop between the software modeling of the E2E model and the accelerator hardware design as shown on the far right side of Figure~\ref{fig:autopilot-high-level}.

The neural layer dimensions of each selected E2E model are configured on the FlexACL programmable accelerator template and its accompanying C++ co-simulator.
The weights of the E2E models are sent from the ML framework to the FlexACL accelerator simulator. 
The computed activations from the accelerator hardware are compared against the software activations to make sure the error difference is within tolerable margins, typically not more than 1e-3, although this margin can be adjusted. 
In case the error target for the activation is met, the FlexACL flow proceeds to the high-level synthesis (HLS) phase. Otherwise, the accelerator design, in its SystemC abstraction, needs to be returned to fix the source of the significant numerical mismatch.

On each accelerator candidate, FlexACL then uses the Catapult HLS tool to auto-generate the RTL from a SystemC source code. During the HLS phase, constraints are set with the goal to achieve maximum throughput on the pipelined design. The accelerator design is further optimized by mapping different memory structures to either SRAMs or latches. 

Finally, FlexACL contains place-and-route utilities to auto-generate a floorplan-aware and timing-aware ASIC layout from the produced HLS RTL. 

%% file: eval.tex
\section{Evaluation}
\label{sec:eval}
In this section, we describe our evaluation methodology for all the components used in the Autopilot infrastructure. First, we describe different settings used to generate environments with varying levels of obstacle density. Then we describe the neural network architecture used as the policy for the DQN algorithm. For the FlexACL framework, we describe the various accelerator parameters such as the number of PEs, memory size, etc. used for accelerator design space exploration. 

\subsection{Training Using Air Learning}
\label{sec:airl-env-settings}

\begin{figure}
        \includegraphics[width=0.85\columnwidth]{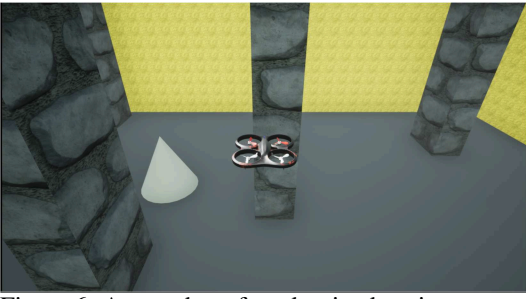}
        \vspace{-10pt}
        \caption{A snapshot of randomized environment generated in Air Learning.}
        \label{fig:env}
\end{figure}

Air Learning has a configurable environment generator\footnote{\url{https://bit.ly/2rlkzy5}} that allows us to change various parameters such as the number of obstacles, size of the arena, seed, etc. We make use of these parameters to generate randomized environments. The environments are generated to increase the complexity of the navigation task for the aerial robot. \Fig{fig:env} shows the snapshot of the generated environments from Air Learning environments.

The specific settings for each of the environments are tabulated in Table~\ref{tab:env}. In this study, we keep the arena size fixed to 25 m $\times$ 25 m $\times$ 20 m. This arena-size is typical and is twice the arena sizes used in aerial robotics testbeds~\cite{flying-arena1,flying-arena2,flying-arena3,flying-arena4}. We also randomly change the seed and goal position in every episode of the training process to improve generalization~\cite{random-goal1,random-goal2}. We also change the seed parameter so that the position of the obstacles is also random. It is shown that randomization is known to improve the generalization of the model to unforeseen situations\cite{domain-rand1}. Since we are only using a depth sensor, we do not randomize textures or other color features available in the Air Learning environment generator.

\begin{table}[]
\centering

\resizebox{0.55\columnwidth}{!}{%

\begin{tabular}{|l|l|}
\hline
\tiny
\textbf{Parameters} & \tiny \textbf{Range} \\ \hline
\tiny \textbf{Arena Size} & \tiny \textit{{[}25m, 25m, 20m{]}} \\ \hline
\tiny \textbf{Static Obstacles} & \tiny \textit{{[}1, 5{]}} \\ \hline
\tiny \textbf{Seed} & \tiny \textit{Random} \\ \hline
\tiny \textbf{Goal Position} & \tiny \textit{Random} \\ \hline
\end{tabular}%
}
\vspace{-10pt}
\caption{Parameters used in the Air Learning environment generator.}
\label{tab:env}
\end{table}

The E2E model is trained using Deep Q-Networks~\cite{dqn}. Prior work has shown that DQN works well on high-level navigation tasks for aerial robots~\cite{dqn-uav1,dqn-uav2}.
The input to the policy is sensor mounted on the drone along with IMU measurements. The output of the policy is one of the 25 actions with different velocity and yaw rates. The reward function we use in this study is defined based on the following equation:

\begin{equation}
r =1000*\alpha - 100*\beta - D_{g} - D_{c}*\delta -1
 \label{eq:reward}
\end{equation}

Here, $\alpha$ is a binary variable whose value is `1' if the agent reaches the goal else its value is `0'.~$\beta$ is a binary variable which is set to `1' if the aerial robot collides with any obstacle or runs out of the maximum allocated steps for an episode.\footnote{We set the maximum allowed steps in an episode as 750. This is to make sure the agent finds the endpoint (goal) within some finite amount of steps.} Otherwise, $\beta$ is '0', effectively penalizing the agent for hitting an obstacle or not reaching the endpoint in time. $D_g$ is the distance to the endpoint from the agent's current location, motivating the agent to move closer to the goal. D$_{c}$ is the distance correction, which is applied to penalize the agent if it chooses actions which speed up the agent away from the goal. The distance correction term is defined as follows:
\begin{equation}
D_{c} = (V_{max} - V_{now})*t_{max}   
\end{equation}

V$_{max}$ is the maximum velocity possible for the agent, which for DQN is fixed at 2.5~$m/s$. V$_{now}$ is the current velocity of the agent, and t$_{max}$ is the duration of the actuation.

\begin{figure}
        \includegraphics[width=0.79\columnwidth]{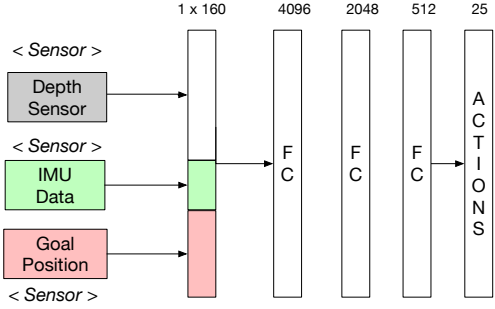}
        \caption{E2E architecture for the policy.}
        \label{fig:network-arch}
\end{figure}

The network architecture used in this work is shown in \Fig{fig:network-arch}. The input comprises of three sets sensors, namely Depth sensor, IMU data such as the velocity of the aerial robot, and distance from the goal position. The sensor inputs are fused into a 1-D array of size 1 x 160. It is then passed to three hidden layers of size 4096, 2048, and 512, respectively. The last FC layer as the same dimension as the action space. The action space for DQN consists of twenty-five discrete actions. Out of these twenty-five action spaces, ten actions are for moving forward with different fixed velocities ranging from 1~$m/s$ to 5~$m/s$, and five actions are for moving backward, five actions for yawing right with fixed yaw rates of 108~\textdegree, 54~\textdegree, 27~\textdegree, 13.5~\textdegree and 6.75~\textdegree  and another five actions for yawing left with fixed yaw rates of -216~\textdegree, -108~\textdegree, -54~\textdegree, -27~\textdegree and -13.5~\textdegree. At each time step, the policy takes observation space as inputs and outputs one of the twenty-five actions based on the observation.

\subsection{FlexACL Design Space} \label{sec:flexacl-temp}

\begin{table}[]
\resizebox{0.75\columnwidth}{!}{%
\begin{tabular}{|l|l|}
\hline
Number of PEs                   & 2, 4, 8, 16, 32 \\ \hline
Number of MAC lanes             & 4, 8, 16        \\ \hline
Weight Buffer Size              & 16kB-1MB        \\ \hline
Weight and Activation Precision & 4-bits, 8-bits  \\ \hline
Input Buffer Size               & 4kB       \\ \hline
Global Buffer Size              & 4kB       \\ \hline
Frequency                       & 300MHz          \\ \hline
\end{tabular}
}
\caption{Design space parameters of the FlexACL accelerator}
\label{tab:flexacl_params}
\end{table}

For vast design space exploration on the FlexACL accelerator template, we vary the design parameters shown in Table~\ref{tab:flexacl_params} in search of an accelerator candidate meeting the desired energy and performance target. Notably, the number of PEs and the number of MAC lanes are swept from 2 to 32 --- and from 4 to 16, respectively, in $2\times$ increment. 
This informs the weight buffer size as accelerators with fewer PEs need a larger scratchpad to store the network's weights. 
The PE's input buffer and GB size are fixed to 4KB, matching the largest activation size of the Airlearning E2E policy, which is 4096, as shown in Figure~\ref{fig:network-arch}. 
The FlexACL accelerator template can also be configured to perform MAC operations at either 8-bit or 4-bit precision for the twice compression.

The performance, power, and area of the generated accelerator candidate are measured on the post-HLS Verilog RTL using a commercial 16nm standard cell library. For this project, the accelerator candidates are synthesized at 300MHz clock frequency, although the clock frequency can be used as another design knob in the energy and performance tradeoff. 

%% file: results.tex
\section{Results}
\label{sec:results}
In this section, we discuss the experimental results based on the evaluation methodology described in Section~\ref{sec:eval}. First, we discuss the performance of the E2E model for the aerial robot task. Then we use the E2E model and generate various accelerator candidates with different performance/power/area trade-offs.

\begin{figure}
        \includegraphics[width=\columnwidth]{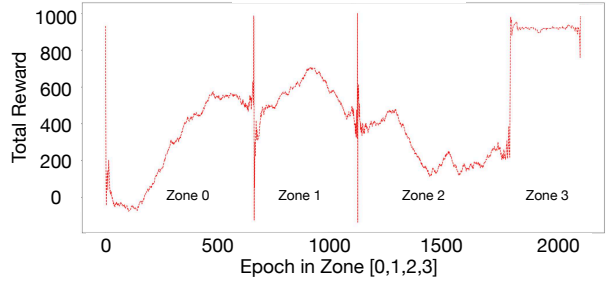}
        \vspace{-20pt}
        \caption{Reward vs epoch in different zones. The sharp drop in reward corresponds to transition to a different zone where the difficulty is higher.}
        \label{fig:reward}
\end{figure}

\begin{figure}
        \includegraphics[width=\columnwidth]{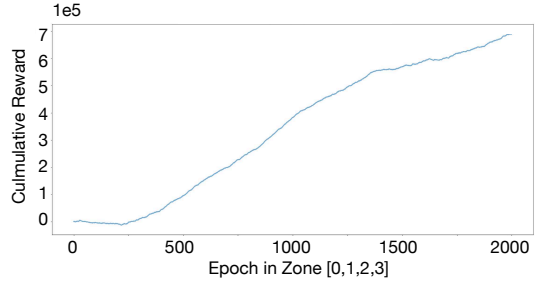}
       \vspace{-20pt} \caption{Cumulative reward for the E2E model over the training duration. An upwatd cumulative reward indicates improvement in agents performance during the training phase.}
        \label{fig:cumulative-reward}
\end{figure}

\subsection{E2E model Performance}

The performance of the DQN model is shown in \Fig{fig:reward}. We observe that agent rewards increases during training and then suddenly drops. This trend continues and finally plateaus at a reward of ~1000 at approximately 2000 epochs.  This trend is expected because, during the training phase, the model is trained using curriculum learning~\cite{bengio2009curriculum}. The arena is divided into multiple zones, and the goal position is randomly placed within that zone. As the agent surpasses a threshold in success rate, the goal position is moved to the next zone. The agent`s performance in each zone is annotated in the \Fig{fig:reward}. The cumulative reward is another metric that quantifies the agent`s performance during the training process~\cite{sutton2018reinforcement}. An increasing cumulative reward signifies that the agent is making progress in reaching the goal position. \Fig{fig:cumulative-reward} shows the cumulative reward for the DQN agent with the E2E policy, as described in \Fig{fig:network-arch}. Once the training stage is complete, the policy is evaluated in a randomized environment for 100 trajectories (episodes). Out of 100 episodes, the agent successfully navigates in a cluttered environment and has a success rate of 91\%.

\begin{figure}[t!]
\centering
\begin{subfigure}{1.0\columnwidth}
\centering
        \includegraphics[width=0.85\columnwidth]{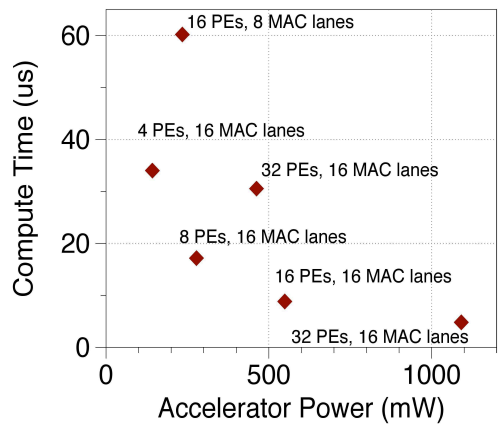}
        \caption{Performance vs Power tradeoffs.}
        \label{fig:flexacl-pow-perf}
        \end{subfigure}
        \begin{subfigure}{1.0\columnwidth}
        \centering
        \includegraphics[width=0.85\columnwidth]{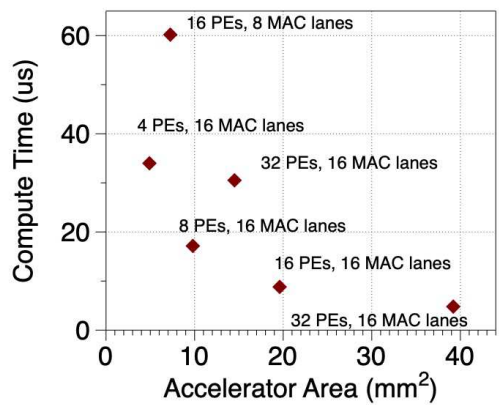}

        \caption{Performance vs Area tradeoffs.}
        \label{fig:flexacl-perf}
        \end{subfigure}
    \label{fig:fund-relation}
 \vspace{-0.05in}
  \caption{(a) Power efficiency of FlexACL accelerator candidates. (b) Area efficiency of FlexACL accelerator candidates. \vspace{-30pt}}

\end{figure}

\begin{figure}
\includegraphics[width=0.8\columnwidth]{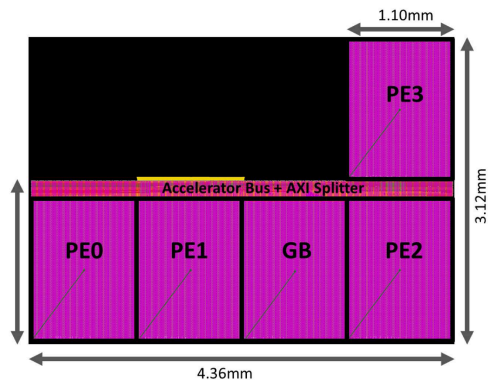}
        \caption{Example placed-and-routed layout of a FlexACL accelerator}
        \label{fig:flexacl_layout}
        \vspace{-20pt}
\end{figure}

\subsection{FlexACL Performance}
Figure~\ref{fig:flexacl-pow-perf}~and~\ref{fig:flexacl-perf} illustrate the performance vs. power and performance vs. area trade-offs, respectively, for the generated FlexACL accelerator candidates. We note that we were not able to take full advantage of all the design space parameters described in Table~\ref{tab:flexacl_params} due to time constraints. Nonetheless, we can observe that an optimized performance, power, and area compromise is distilled in the 8PEs-16MAC lanes accelerator as it represents the knee of the lower-left curve in both figures.
The 4PEs-16MAC lanes accelerator consumes the smallest power and area while, not surprisingly, the 32PEs-16MAC lanes accelerator consumes the highest power and area but yields the shortest latency.

Figure~\ref{fig:flexacl_layout} shows an example placed-and-routed layout of an accelerator with 4 PEs, each with 16 MAC lanes. 

%% file: conclusion.tex
\section{Conclusion}
\label{sec:conclusion}
In this paper, we present \autop framework to automate algorithm-accelerator co-design for aerial robots. We demonstrated the efficacy of the AutoSoC framework by designing an E2E model for autonomous navigation tasks in a cluttered environment and generate various accelerator candidates to process the E2E model efficiently. The E2E model we designed has a success rate of 91\% for the given robot task. The generated accelerator candidates have the performance range from 4.8 microseconds to 60.2 microseconds with a power envelope of 0.142 W ~to 1.09 W. The ability to generate a variety of hardware accelerators with different performance and power budgets allows targeting different categories of the aerial robot. Also, the methodology and key infrastructure components we have are generic and can be applied to other autonomous machines. 

\section{Acknowledgements}
This work was carried as a course project for CS249r Special Topics on Edge Computing (Autonomous Machines) at Harvard University. We thank Prof. Vijay Janapa Reddi for constructive feedback and discussions.

%% file: main.bbl
\begin{thebibliography}{70}
\providecommand{\natexlab}[1]{#1}
\providecommand{\url}[1]{\texttt{#1}}
\expandafter\ifx\csname urlstyle\endcsname\relax
  \providecommand{\doi}[1]{doi: #1}\else
  \providecommand{\doi}{doi: \begingroup \urlstyle{rm}\Url}\fi

\bibitem[20m(2019)]{20mins}
The top 5 fpv racing drones: Ready-to-fly models for drone racing, 2019.
\newblock URL \url{https://uavcoach.com/fpv-quadcopter-drone-systems/}.

\bibitem[Abadi et~al.(2016)Abadi, Barham, Chen, Chen, Davis, Dean, Devin,
  Ghemawat, Irving, Isard, et~al.]{abadi2016tensorflow}
Abadi, M., Barham, P., Chen, J., Chen, Z., Davis, A., Dean, J., Devin, M.,
  Ghemawat, S., Irving, G., Isard, M., et~al.
\newblock Tensorflow: A system for large-scale machine learning.
\newblock In \emph{12th $\{$USENIX$\}$ Symposium on Operating Systems Design
  and Implementation ($\{$OSDI$\}$ 16)}, pp.\  265--283, 2016.

\bibitem[Arm()]{axi}
Arm.
\newblock Introduction to axi protocol: Understanding the axi interface.
\newblock
  \url{https://community.arm.com/developer/ip-products/system/b/soc-design-blog/posts/introduction-to-axi-protocol-understanding-the-axi-interface}.

\bibitem[{Bacco} et~al.(2018){Bacco}, {Berton}, {Ferro}, {Gennaro}, {Gotta},
  {Matteoli}, {Paonessa}, {Ruggeri}, {Virone}, and {Zanella}]{8373043}
{Bacco}, M., {Berton}, A., {Ferro}, E., {Gennaro}, C., {Gotta}, A., {Matteoli},
  S., {Paonessa}, F., {Ruggeri}, M., {Virone}, G., and {Zanella}, A.
\newblock Smart farming: Opportunities, challenges and technology enablers.
\newblock In \emph{2018 IoT Vertical and Topical Summit on Agriculture -
  Tuscany (IOT Tuscany)}, pp.\  1--6, May 2018.
\newblock \doi{10.1109/IOT-TUSCANY.2018.8373043}.

\bibitem[Bengio et~al.(2009)Bengio, Louradour, Collobert, and
  Weston]{bengio2009curriculum}
Bengio, Y., Louradour, J., Collobert, R., and Weston, J.
\newblock Curriculum learning.
\newblock In \emph{Proceedings of the 26th annual international conference on
  machine learning}, pp.\  41--48. ACM, 2009.

\bibitem[Bojarski et~al.(2016)Bojarski, Del~Testa, Dworakowski, Firner, Flepp,
  Goyal, Jackel, Monfort, Muller, Zhang, et~al.]{e2e-nvidia}
Bojarski, M., Del~Testa, D., Dworakowski, D., Firner, B., Flepp, B., Goyal, P.,
  Jackel, L.~D., Monfort, M., Muller, U., Zhang, J., et~al.
\newblock End to end learning for self-driving cars.
\newblock \emph{arXiv preprint arXiv:1604.07316}, 2016.

\bibitem[Boroujerdian et~al.(2018)Boroujerdian, Genc, Krishnan, Cui, Almeida,
  Mansoorshahi, Faust, and Reddi]{Bor2018}
Boroujerdian, B., Genc, H., Krishnan, S., Cui, W., Almeida, M., Mansoorshahi,
  K., Faust, A., and Reddi, V.~J.
\newblock Mavbench: Micro aerial vehicle benchmarking.
\newblock In \emph{51st Annual IEEE/ACM International Symposium on
  Microarchitecture (MICRO)}, pp.\  894--907, 2018.

\bibitem[Brown(2017)]{1}
Brown, J.
\newblock Estes 4606 proto x nano drone: Specs, features, reviews, prices,
  competitors, Jun 2017.
\newblock URL
  \url{https://www.mydronelab.com/reviews/estes-4606-proto-x-nano.html}.

\bibitem[Chen et~al.(2014)Chen, Du, Sun, Wang, Wu, Chen, and Temam]{diannao}
Chen, T., Du, Z., Sun, N., Wang, J., Wu, C., Chen, Y., and Temam, O.
\newblock Diannao: A small-footprint high-throughput accelerator for ubiquitous
  machine-learning.
\newblock In \emph{Proceedings of the 19th International Conference on
  Architectural Support for Programming Languages and Operating Systems},
  ASPLOS '14, pp.\  269--284, New York, NY, USA, 2014. ACM.

\bibitem[{Chen} et~al.(2016){Chen}, {Emer}, and {Sze}]{eyeriss}
{Chen}, Y., {Emer}, J., and {Sze}, V.
\newblock Eyeriss: A spatial architecture for energy-efficient dataflow for
  convolutional neural networks.
\newblock In \emph{2016 ACM/IEEE 43rd Annual International Symposium on
  Computer Architecture (ISCA)}, pp.\  367--379, June 2016.

\bibitem[Conroy et~al.(2009)Conroy, Gremillion, Ranganathan, and
  Humbert]{Conroy:2009:IWI:1644233.1644234}
Conroy, J., Gremillion, G., Ranganathan, B., and Humbert, J.~S.
\newblock Implementation of wide-field integration of optic flow for autonomous
  quadrotor navigation.
\newblock \emph{Auton. Robots}, 27\penalty0 (3):\penalty0 189--198, October
  2009.
\newblock ISSN 0929-5593.
\newblock \doi{10.1007/s10514-009-9140-0}.
\newblock URL \url{http://dx.doi.org/10.1007/s10514-009-9140-0}.

\bibitem[Da-Yu et~al.(2019)Da-Yu, Min-Ching, Wen-Ying, Jsen-Shung, Chien-Hung,
  and Fuching]{spark}
Da-Yu, K., Min-Ching, C., Wen-Ying, W., Jsen-Shung, L., Chien-Hung, C., and
  Fuching, T.
\newblock Drone forensic investigation: Dji spark drone as a case study.
\newblock \emph{Procedia Computer Science}, 159:\penalty0 1890--1899, 2019.

\bibitem[Feltman()]{photographyh-org}
Feltman, R.
\newblock The future of sports photography:drones.
\newblock \url{https://www.theatlantic.com/technology/archive/2014/02/ the-
  future- of- sports- photography- drones/283896/}.

\bibitem[Finn et~al.(2017)Finn, Abbeel, and Levine]{random-goal2}
Finn, C., Abbeel, P., and Levine, S.
\newblock Model-agnostic meta-learning for fast adaptation of deep networks.
\newblock In \emph{Proceedings of the 34th International Conference on Machine
  Learning-Volume 70}, pp.\  1126--1135. JMLR. org, 2017.

\bibitem[Gandhi et~al.(2017)Gandhi, Pinto, and Gupta]{learn2crash}
Gandhi, D., Pinto, L., and Gupta, A.
\newblock Learning to fly by crashing.
\newblock In \emph{2017 IEEE/RSJ International Conference on Intelligent Robots
  and Systems (IROS)}, pp.\  3948--3955. IEEE, 2017.

\bibitem[Gang()]{entertainment-org}
Gang, J.
\newblock Drone use in the entertainment industry and beyond.
\newblock
  \url{https://thebottomline.as.ucsb.edu/2018/09/drone-use-in-the-entertainment-industry-and-beyond}.

\bibitem[Han et~al.(2016)Han, Liu, Mao, Pu, Pedram, Horowitz, and Dally]{eie}
Han, S., Liu, X., Mao, H., Pu, J., Pedram, A., Horowitz, M.~A., and Dally,
  W.~J.
\newblock Eie: Efficient inference engine on compressed deep neural network.
\newblock \emph{SIGARCH Comput. Archit. News}, 44\penalty0 (3), June 2016.

\bibitem[Hart et~al.(1968)Hart, Nilsson, and Raphael]{a-star}
Hart, P.~E., Nilsson, N.~J., and Raphael, B.
\newblock A formal basis for the heuristic determination of minimum cost paths.
\newblock \emph{IEEE transactions on Systems Science and Cybernetics},
  4\penalty0 (2):\penalty0 100--107, 1968.

\bibitem[Hill et~al.(2018)Hill, Raffin, Ernestus, Gleave, Kanervisto, Traore,
  Dhariwal, Hesse, Klimov, Nichol, Plappert, Radford, Schulman, Sidor, and
  Wu]{stable-baselines}
Hill, A., Raffin, A., Ernestus, M., Gleave, A., Kanervisto, A., Traore, R.,
  Dhariwal, P., Hesse, C., Klimov, O., Nichol, A., Plappert, M., Radford, A.,
  Schulman, J., Sidor, S., and Wu, Y.
\newblock Stable baselines.
\newblock \url{https://github.com/hill-a/stable-baselines}, 2018.

\bibitem[Holly(2019)]{falcore}
Holly, H.
\newblock Drone review: Connex falcore, Apr 2019.
\newblock URL
  \url{https://www.rotordronepro.com/drone-review-connex-falcore/#outer-popup}.

\bibitem[Hornung et~al.(2013)Hornung, Wurm, Bennewitz, Stachniss, and
  Burgard]{octomap}
Hornung, A., Wurm, K.~M., Bennewitz, M., Stachniss, C., and Burgard, W.
\newblock Octomap: An efficient probabilistic 3d mapping framework based on
  octrees.
\newblock \emph{Autonomous robots}, 34\penalty0 (3):\penalty0 189--206, 2013.

\bibitem[How et~al.()How, Teo, and Michini]{flying-arena3}
How, J.~P., Teo, J., and Michini, B.
\newblock Adaptive flight control experiments using raven.
\newblock \emph{Simulation}, 1:\penalty0 1.

\bibitem[Kalashnikov et~al.(2018)Kalashnikov, Irpan, Pastor, Ibarz, Herzog,
  Jang, Quillen, Holly, Kalakrishnan, Vanhoucke, et~al.]{qt-opt}
Kalashnikov, D., Irpan, A., Pastor, P., Ibarz, J., Herzog, A., Jang, E.,
  Quillen, D., Holly, E., Kalakrishnan, M., Vanhoucke, V., et~al.
\newblock Qt-opt: Scalable deep reinforcement learning for vision-based robotic
  manipulation.
\newblock \emph{arXiv preprint arXiv:1806.10293}, 2018.

\bibitem[Kavraki et~al.(1996)Kavraki, Svestka, Latombe, and Overmars]{prm}
Kavraki, L.~E., Svestka, P., Latombe, J.-C., and Overmars, M.~H.
\newblock Probabilistic roadmaps for path planning in high-dimensional
  configuration spaces.
\newblock \emph{IEEE transactions on Robotics and Automation}, 12\penalty0
  (4):\penalty0 566--580, 1996.

\bibitem[Kendall et~al.(2019)Kendall, Hawke, Janz, Mazur, Reda, Allen, Lam,
  Bewley, and Shah]{rl-car}
Kendall, A., Hawke, J., Janz, D., Mazur, P., Reda, D., Allen, J.-M., Lam,
  V.-D., Bewley, A., and Shah, A.
\newblock Learning to drive in a day.
\newblock In \emph{2019 International Conference on Robotics and Automation
  (ICRA)}, pp.\  8248--8254. IEEE, 2019.

\bibitem[Krishnan et~al.(2019)Krishnan, Boroujerdian, Fu, Faust, and
  Reddi]{krishnan2019}
Krishnan, S., Boroujerdian, B., Fu, W., Faust, A., and Reddi, V.~J.
\newblock Air learning: An {AI} research platform for algorithm-hardware
  benchmarking of autonomous aerial robots.
\newblock \emph{arXiv:1906.00421}, 2019.

\bibitem[Krishnan et~al.(2021)Krishnan, Boroujerdian, Fu, Faust, and
  Reddi]{airlearning}
Krishnan, S., Boroujerdian, B., Fu, W., Faust, A., and Reddi, V.~J.
\newblock Air learning: a deep reinforcement learning gym for autonomous aerial
  robot visual navigation.
\newblock \emph{Machine Learning}, pp.\  1--40, 2021.

\bibitem[LaValle(1998)]{rrt}
LaValle, S.~M.
\newblock Rapidly-exploring random trees: A new tool for path planning.
\newblock 1998.

\bibitem[Lee \& Choi(2016)Lee and Choi]{mining}
Lee, S. and Choi, Y.
\newblock Reviews of unmanned aerial vehicle (drone) technology trends and its
  applications in the mining industry.
\newblock \emph{Geosystem Engineering}, 19\penalty0 (4):\penalty0 197--204,
  2016.
\newblock \doi{10.1080/12269328.2016.1162115}.
\newblock URL \url{https://doi.org/10.1080/12269328.2016.1162115}.

\bibitem[Loquercio et~al.(2018)Loquercio, Maqueda, Del-Blanco, and
  Scaramuzza]{dronet}
Loquercio, A., Maqueda, A.~I., Del-Blanco, C.~R., and Scaramuzza, D.
\newblock Dronet: Learning to fly by driving.
\newblock \emph{IEEE Robotics and Automation Letters}, 3\penalty0 (2):\penalty0
  1088--1095, 2018.

\bibitem[Lupashin et~al.(2014)Lupashin, Hehn, Mueller, Schoellig, Sherback, and
  D’Andrea]{flying-arena1}
Lupashin, S., Hehn, M., Mueller, M.~W., Schoellig, A.~P., Sherback, M., and
  D’Andrea, R.
\newblock A platform for aerial robotics research and demonstration: The flying
  machine arena.
\newblock \emph{Mechatronics}, 24\penalty0 (1):\penalty0 41--54, 2014.

\bibitem[Max \& Steven(2018)Max and Steven]{holy}
Max, M. and Steven, T.
\newblock Holy stone hs170 predator review, Nov 2018.
\newblock URL
  \url{https://www.techgearlab.com/reviews/cool-gadgets/drones-under-100/holy-stone-hs170-predator}.

\bibitem[McCullough()]{surveillance-org}
McCullough, D. R.~C.
\newblock Unmanned aircraft systems(uas) guidebook in development.
\newblock \url{https://cops.usdoj.gov/html/dispatch/ 08-2014/UAS Guidebook in
  Development.asp}.

\bibitem[{McGuire} et~al.(2017){McGuire}, {de Croon}, {De Wagter}, {Tuyls}, and
  {Kappen}]{7833065}
{McGuire}, K., {de Croon}, G., {De Wagter}, C., {Tuyls}, K., and {Kappen}, H.
\newblock Efficient optical flow and stereo vision for velocity estimation and
  obstacle avoidance on an autonomous pocket drone.
\newblock \emph{IEEE Robotics and Automation Letters}, 2\penalty0 (2):\penalty0
  1070--1076, April 2017.
\newblock ISSN 2377-3774.
\newblock \doi{10.1109/LRA.2017.2658940}.

\bibitem[Michael et~al.(2010)Michael, Mellinger, Lindsey, and
  Kumar]{flying-arena2}
Michael, N., Mellinger, D., Lindsey, Q., and Kumar, V.
\newblock The grasp multiple micro-uav testbed.
\newblock \emph{IEEE Robotics \& Automation Magazine}, 17\penalty0
  (3):\penalty0 56--65, 2010.

\bibitem[Michaels()]{survey-org}
Michaels, M.
\newblock Drones: Helping noaa from hurricanes to red tide.
\newblock
  \url{https://www.weathernationtv.com/news/drones-helping-noaa-from-hurricanes-to-red-tide/}.

\bibitem[Michel()]{package-org-2}
Michel, H.
\newblock Amazon's drone patents.
\newblock \url{ http://dronecenter. bard.edu/amazon- drone- patents/}.

\bibitem[Migacz(2017)]{migacz2017}
Migacz, S.
\newblock 8-bit inference with tensorrt.
\newblock In \emph{NVIDIA GPU Tech Conf}, 2017.
\newblock URL
  \url{http://on-demand.gputechconf.com/gtc/2017/presentation/s7310-8-bit-inference-with-tensorrt.pdf}.

\bibitem[Mnih et~al.(2013)Mnih, Kavukcuoglu, Silver, Graves, Antonoglou,
  Wierstra, and Riedmiller]{dqn}
Mnih, V., Kavukcuoglu, K., Silver, D., Graves, A., Antonoglou, I., Wierstra,
  D., and Riedmiller, M.
\newblock Playing atari with deep reinforcement learning.
\newblock \emph{arXiv preprint arXiv:1312.5602}, 2013.

\bibitem[Mohta et~al.(2018)Mohta, Watterson, Mulgaonkar, Liu, Qu, Makineni,
  Saulnier, Sun, Zhu, Delmerico, et~al.]{high-speed-ppc}
Mohta, K., Watterson, M., Mulgaonkar, Y., Liu, S., Qu, C., Makineni, A.,
  Saulnier, K., Sun, K., Zhu, A., Delmerico, J., et~al.
\newblock Fast, autonomous flight in gps-denied and cluttered environments.
\newblock \emph{Journal of Field Robotics}, 35\penalty0 (1):\penalty0 101--120,
  2018.

\bibitem[Momont()]{medical-org}
Momont, A.
\newblock Ambulance drone.
\newblock \url{https: //www.tudelft.nl/en/ide/research/research-labs/applied-
  labs/ambulance-drone/}.

\bibitem[Mur-Artal \& Tard{\'o}s(2017)Mur-Artal and Tard{\'o}s]{slam}
Mur-Artal, R. and Tard{\'o}s, J.~D.
\newblock Orb-slam2: An open-source slam system for monocular, stereo, and
  rgb-d cameras.
\newblock \emph{IEEE Transactions on Robotics}, 33\penalty0 (5):\penalty0
  1255--1262, 2017.

\bibitem[Packer et~al.(2018)Packer, Gao, Kos, Kr{\"a}henb{\"u}hl, Koltun, and
  Song]{random-goal1}
Packer, C., Gao, K., Kos, J., Kr{\"a}henb{\"u}hl, P., Koltun, V., and Song, D.
\newblock Assessing generalization in deep reinforcement learning.
\newblock \emph{arXiv preprint arXiv:1810.12282}, 2018.

\bibitem[Palossi et~al.(2018)Palossi, Loquercio, Conti, Flamand, Scaramuzza,
  and Benini]{DBLP:journals/corr/abs-1805-01831}
Palossi, D., Loquercio, A., Conti, F., Flamand, E., Scaramuzza, D., and Benini,
  L.
\newblock Ultra low power deep-learning-powered autonomous nano drones.
\newblock \emph{CoRR}, abs/1805.01831, 2018.
\newblock URL \url{http://arxiv.org/abs/1805.01831}.

\bibitem[Palossi et~al.(2019)Palossi, Loquercio, Conti, Flamand, Scaramuzza,
  and Benini]{pulp-dronet}
Palossi, D., Loquercio, A., Conti, F., Flamand, E., Scaramuzza, D., and Benini,
  L.
\newblock A 64mw dnn-based visual navigation engine for autonomous nano-drones.
\newblock \emph{IEEE Internet of Things Journal}, 2019.

\bibitem[Palunko et~al.(2012)Palunko, Cruz, and Fierro]{flying-arena4}
Palunko, I., Cruz, P., and Fierro, R.
\newblock Agile load transportation: Safe and efficient load manipulation with
  aerial robots.
\newblock \emph{IEEE robotics \& automation magazine}, 19\penalty0
  (3):\penalty0 69--79, 2012.

\bibitem[Polvara et~al.(2018)Polvara, Patacchiola, Sharma, Wan, Manning,
  Sutton, and Cangelosi]{dqn-uav1}
Polvara, R., Patacchiola, M., Sharma, S., Wan, J., Manning, A., Sutton, R., and
  Cangelosi, A.
\newblock Toward end-to-end control for uav autonomous landing via deep
  reinforcement learning.
\newblock In \emph{2018 International Conference on Unmanned Aircraft Systems
  (ICUAS)}, pp.\  115--123. IEEE, 2018.

\bibitem[Qiantori et~al.(2012)Qiantori, Sutiono, Hariyanto, Suwa, and
  Ohta]{search-and-rescue}
Qiantori, A., Sutiono, A.~B., Hariyanto, H., Suwa, H., and Ohta, T.
\newblock An emergency medical communications system by low altitude platform
  at the early stages of a natural disaster in indonesia.
\newblock \emph{J. Med. Syst.}, 36, 2012.

\bibitem[Qin et~al.(2018)Qin, Li, and Shen]{vins-mono}
Qin, T., Li, P., and Shen, S.
\newblock Vins-mono: A robust and versatile monocular visual-inertial state
  estimator.
\newblock \emph{IEEE Transactions on Robotics}, 34\penalty0 (4):\penalty0
  1004--1020, 2018.

\bibitem[Reagen et~al.(2016)Reagen, Whatmough, Adolf, Rama, Lee, Lee,
  Hern\'{a}ndez-Lobato, Wei, and Brooks]{minerva}
Reagen, B., Whatmough, P., Adolf, R., Rama, S., Lee, H., Lee, S.~K.,
  Hern\'{a}ndez-Lobato, J.~M., Wei, G.-Y., and Brooks, D.
\newblock Minerva: Enabling low-power, highly-accurate deep neural network
  accelerators.
\newblock In \emph{Proceedings of the 43rd International Symposium on Computer
  Architecture}, ISCA '16, pp.\  267--278, Piscataway, NJ, USA, 2016.
\newblock ISBN 978-1-4673-8947-1.

\bibitem[Richard(2016)]{typhoon}
Richard, B.
\newblock Yuneec typhoon h drone review, Aug 2016.
\newblock URL
  \url{https://www.tomsguide.com/us/yuneec-typhoon-h-drone,review-3858.html}.

\bibitem[Rick(2018)]{6mins}
Rick, P.
\newblock 12 drones with long flight time, Jan 2018.
\newblock URL \url{http://www.dronesfella.com/guide/long-flight-time/}.

\bibitem[Rogers()]{search-and-rescue-org}
Rogers, J.
\newblock How drones are helping the nepal earthquake relief effort.
\newblock \url{http://www.foxnews.com/tech/2015/04/30/ how- drones- are-
  helping- nepal- earthquake- relief- effort.html}.

\bibitem[Ross et~al.(2013)Ross, Melik-Barkhudarov, Shankar, Wendel, Dey,
  Bagnell, and Hebert]{trailnet0}
Ross, S., Melik-Barkhudarov, N., Shankar, K.~S., Wendel, A., Dey, D., Bagnell,
  J.~A., and Hebert, M.
\newblock Learning monocular reactive uav control in cluttered natural
  environments.
\newblock In \emph{2013 IEEE international conference on robotics and
  automation}, pp.\  1765--1772. IEEE, 2013.

\bibitem[Sadeghi \& Levine(2016)Sadeghi and Levine]{cad2rl}
Sadeghi, F. and Levine, S.
\newblock Cad2rl: Real single-image flight without a single real image.
\newblock \emph{arXiv preprint arXiv:1611.04201}, 2016.

\bibitem[Smolyanskiy et~al.(2017{\natexlab{a}})Smolyanskiy, Kamenev, Smith, and
  Birchfield]{trail-net}
Smolyanskiy, N., Kamenev, A., Smith, J., and Birchfield, S.
\newblock Toward low-flying autonomous mav trail navigation using deep neural
  networks for environmental awareness.
\newblock In \emph{2017 IEEE/RSJ International Conference on Intelligent Robots
  and Systems (IROS)}, pp.\  4241--4247. IEEE, 2017{\natexlab{a}}.

\bibitem[Smolyanskiy et~al.(2017{\natexlab{b}})Smolyanskiy, Kamenev, Smith, and
  Birchfield]{trailnet}
Smolyanskiy, N., Kamenev, A., Smith, J., and Birchfield, S.
\newblock Toward low-flying autonomous {MAV} trail navigation using deep neural
  networks for environmental awareness.
\newblock In \emph{IEEE/RSJ International Conference on Intelligent Robots and
  Systems (IROS)}, pp.\  4241--4247, 2017{\natexlab{b}}.

\bibitem[{Suciu} et~al.(2018){Suciu}, {Dragu}, {Hussain}, {Iliescu}, {Orza},
  and {Mocanu}]{8588849}
{Suciu}, G., {Dragu}, M., {Hussain}, I., {Iliescu}, A., {Orza}, O., and
  {Mocanu}, C.
\newblock 3d modeling using parrot bebop 2 fpv.
\newblock In \emph{2018 IEEE 16th International Conference on Embedded and
  Ubiquitous Computing (EUC)}, pp.\  61--65, Oct 2018.
\newblock \doi{10.1109/EUC.2018.00016}.

\bibitem[Sutton \& Barto(2018{\natexlab{a}})Sutton and Barto]{rl-book}
Sutton, R.~S. and Barto, A.~G.
\newblock \emph{Reinforcement learning: An introduction}.
\newblock MIT press, 2018{\natexlab{a}}.

\bibitem[Sutton \& Barto(2018{\natexlab{b}})Sutton and
  Barto]{sutton2018reinforcement}
Sutton, R.~S. and Barto, A.~G.
\newblock \emph{Reinforcement learning: An introduction}.
\newblock 2018{\natexlab{b}}.

\bibitem[Timothy et~al.(2017)Timothy, Paul, Aaron, Joan, and Jeff]{Timothy2017}
Timothy, A., Paul, M.~N., Aaron, A.~T., Joan, B., and Jeff, S.
\newblock Drone transportation of blood products.
\newblock \emph{TRANSFUSION Journal}, 57\penalty0 (3):\penalty0 582--588, 2017.

\bibitem[Tobin et~al.(2017)Tobin, Fong, Ray, Schneider, Zaremba, and
  Abbeel]{domain-rand1}
Tobin, J., Fong, R., Ray, A., Schneider, J., Zaremba, W., and Abbeel, P.
\newblock Domain randomization for transferring deep neural networks from
  simulation to the real world.
\newblock In \emph{2017 IEEE/RSJ International Conference on Intelligent Robots
  and Systems (IROS)}, pp.\  23--30. IEEE, 2017.

\bibitem[Venkataramani et~al.(2017)Venkataramani, Ranjan, Banerjee, Das,
  Avancha, Jagannathan, Durg, Nagaraj, Kaul, Dubey, and Raghunathan]{scaledeep}
Venkataramani, S., Ranjan, A., Banerjee, S., Das, D., Avancha, S., Jagannathan,
  A., Durg, A., Nagaraj, D., Kaul, B., Dubey, P., and Raghunathan, A.
\newblock Scaledeep: A scalable compute architecture for learning and
  evaluating deep networks.
\newblock \emph{SIGARCH Comput. Archit. News}, 2017.

\bibitem[Venkatesan et~al.(2019)]{magnet}
Venkatesan, B. et~al.
\newblock Magnet : A modular accelerator generator for neural networks.
\newblock In \emph{ICCAD}, 2019.

\bibitem[Waibel et~al.(2017)Waibel, Keays, and Augugliaro]{droneshow}
Waibel, M., Keays, B., and Augugliaro, F.
\newblock Drone shows: Creative potential and best practices.
\newblock Technical report, 2017.
\newblock .

\bibitem[Weise()]{package-org-1}
Weise, E.
\newblock Amazon delivered its first customer package by drone.
\newblock \url{https://www.usatoday.com/story/tech/news/2016/12/14/ amazon-
  delivered- its- first- customer- package- drone/95401366/}.

\bibitem[Will(2013)]{parrot}
Will, G.
\newblock Parrot ar.drone 2.0, Sept 2013.
\newblock URL \url{https://www.pcmag.com/review/315789/parrot-ar-drone-2-0}.

\bibitem[Wood et~al.(2012)Wood, Finio, Karpelson, Ma, P{\'e}rez-Arancibia,
  Sreetharan, Tanaka, and Whitney]{Wood:2012:PPA:2370704.2370708}
Wood, R., Finio, B., Karpelson, M., Ma, K., P{\'e}rez-Arancibia, N.,
  Sreetharan, P., Tanaka, H., and Whitney, J.
\newblock Progress on 'pico' air vehicles.
\newblock \emph{Int. J. Rob. Res.}, 31\penalty0 (11):\penalty0 1292--1302,
  September 2012.
\newblock ISSN 0278-3649.
\newblock \doi{10.1177/0278364912455073}.
\newblock URL \url{http://dx.doi.org/10.1177/0278364912455073}.

\bibitem[Yan et~al.(2019)Yan, Xiang, and Wang]{dqn-uav2}
Yan, C., Xiang, X., and Wang, C.
\newblock Towards real-time path planning through deep reinforcement learning
  for a uav in dynamic environments.
\newblock \emph{Journal of Intelligent {\&} Robotic Systems}, Sep 2019.
\newblock ISSN 1573-0409.
\newblock \doi{10.1007/s10846-019-01073-3}.
\newblock URL \url{https://doi.org/10.1007/s10846-019-01073-3}.

\bibitem[{Yang} et~al.(2018){Yang}, {Zheng}, {Bian}, {Song}, and
  {Han}]{8119942}
{Yang}, Y., {Zheng}, Z., {Bian}, K., {Song}, L., and {Han}, Z.
\newblock Real-time profiling of fine-grained air quality index distribution
  using uav sensing.
\newblock \emph{IEEE Internet of Things Journal}, 5\penalty0 (1):\penalty0
  186--198, Feb 2018.
\newblock ISSN 2372-2541.
\newblock \doi{10.1109/JIOT.2017.2777820}.

\end{thebibliography}
